\documentclass[12pt]{ceurart}

\usepackage{microtype}
\usepackage{graphicx}
\usepackage[justification=centering]{caption}
\usepackage{subcaption}
\usepackage{tabularx}
\begin{document}

\copyrightyear{2023}
\copyrightclause{Copyright for this paper by its authors.
  Use permitted under Creative Commons License Attribution 4.0
  International (CC BY 4.0).}

\conference{SAIL'23: 3rd Symposium on Artificial Intelligence and Law, 24-26 February, 2023, Hybrid Event, Hyderabad, India}

\title{Similar Cases Recommendation using Legal Knowledge Graphs}

\author[1,3]{Jaspreet Singh Dhani}[%
email=jsdhani@cs.toronto.edu,
]
\address[1]{University of Delhi, New Delhi, India}

\author[1]{Ruchika Bhatt}[%
email=ruchika.mcs19.du@gmail.com,
]

\author[2]{Balaji Ganesan}[%
email=bganesa1@in.ibm.com,
]
\address[2]{IBM Research, Bengaluru, India}

\author[1]{Parikshet Sirohi}[%
email=parikshet.sirohi@gmail.com,
]

\author[1]{Vasudha Bhatnagar}[%
email=vbhatnagar@cs.du.ac.in,
]
\address[3]{University of Toronto, Toronto, Canada}

\begin{abstract}
A legal knowledge graph constructed from court cases, judgments, laws and other legal documents can enable a number of applications like question answering, document similarity, and search. While the use of knowledge graphs for distant supervision in NLP tasks is well researched, using knowledge graphs for applications like case similarity presents challenges. In this work, we describe our solution for predicting similar cases in Indian court judgements. We present our results and also discuss the impact of large language models on this task.
\end{abstract}

\begin{keywords}
  Legal Knowledge Graphs \sep
  Graph Neural Networks \sep
  Link Prediction \sep
  Large Language Models
\end{keywords}

\maketitle

\section{Introduction}
\label{sec:introduction}

In many countries, the legal system is overwhelmed by a backlog of cases, especially in the lower judiciary. While there are legislation like  speedy justice acts, the legal processes are inherently time consuming. AI tools can help automate some of these processes and speed up justice delivery.

Representing cases filed in courts of law as nodes and citations as edges  in a graph, could enable several graph tasks like link prediction, node similarity and node classification. This representation has the potential to improve further downstream applications of legal document analysis such as legal cognitive assistance \cite{joshi2016alda}, question-answering, text summarization, judgement prediction \cite{zhong-etal-2020-nlp} and finding similar cases.

\begin{figure*}[!htb]
    \centering
    \includegraphics[width=\textwidth]{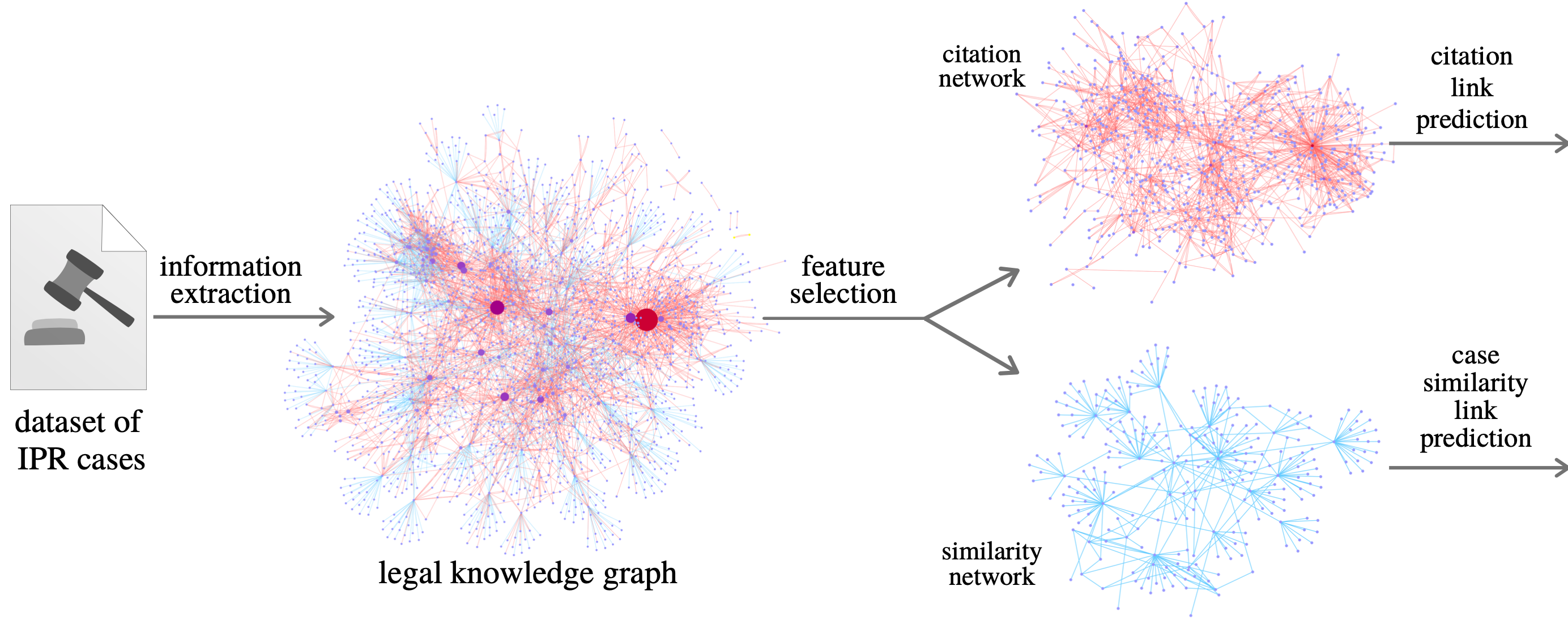}
    \captionsetup{justification=centering}
    \caption{Case similarity and citation link prediction on our case graphs}
    \label{pipeline}
\end{figure*}

We present a case similarity solution using Graph Neural Networks (GNNs), that can help law practitioners to find similar cases that could lead to early settlements, better case documents, and faster judgements. The underlying legal knowledge graph can help students familiarize with the legal terms and concepts. Such knowledge graphs can also be used to infuse knowledge into or fine tune large language models (LLMs) to fill gaps in such models where they may not have sufficient domain specific knowledge.

We approach the problem of node similarity in a case graph, by first constructing a legal knowledge graph using the methods described in \cite{agarwal2017cognitive} and \cite{vannur2021data}. In this work, the legal knowledge graph is used to infuse domain knowledge into downstream tasks. We have used a corpus of court documents, laws, and judgements relating to intellectual property rights (IPR) cases in Indian courts. The nodes in the legal knowledge graph are cases, judicial orders, legislation, people among other entities. Edges are relationships like citations, people involved in a case, sections of laws related to a case among others.

We construct the above legal knowledge graph from unstructured text using a combination of rule-based and graph neural models adhering to an ontology. We then use topic modelling using Latent Dirichlet Allocation (LDA) to surface important concepts to for our ontology and also for feature selection in the citation link prediction and case similarity tasks.

We then train graph neural network models for both the citation prediction and case similarity tasks. We use a vanilla RGCN \cite{schlichtkrull2018rgcn} model on the graph described above. This is our baseline. We then use law points identified by legal experts as handcrafted features in addition to the features in our baseline model. We compare these two models with a third variant which uses LegalBERT \cite{chalkidis2020legal} to encode the node features.

Our contributions are as follows. In Section \ref{legalkg}, we describe a method to construct a legal knowledge graph from Indian court judgements. In Section \ref{sec:case_graph}, we use topic modelling to select features to train graph neural network models for finding similar cases. In Section \ref{sec:experiments}, we conduct experiments comparing this vanilla approach with handcrafted features and LegalBERT and share our results. In Section \ref{sec:discussion}, we include a discussion on alternative approaches to the graph neural networks model that we have developed. And finally in Section \ref{sec:deployment}, we describe how this system is deployed.

\section{Related Work}
\label{sec:related_work}

Constructing a knowledge graph from unstructured documents has been explored in several works. \cite{chiticariu2010systemt} proposed a system to extract domain specific entities and relationships from documents. \cite{agarwal2017cognitive} proposed a platform for creating knowledge graphs for Regulatory Compliance.  \cite{vannur2021data} discussed fairness in personal knowledge base construction. \cite{ganesan2020anu} presented a question answering system that leverages domain specific knowledge graphs. We can characterize all the methods as based on rule based or rules assisted knowledge base construction.

To construct legal knowledge graphs on corpora like judgements, there is a need for some level of customization of the knowledge graph construction pipeline. To begin with extracting and classifying legal entities could be accomplished with fine grained entity classification. \cite{abhishek2019collective} proposed a unified hierarchy to learn entity types. \cite{dasgupta2018fine} discussed fine grained classification of personal data entities. We observe that personal data entities are seldom useful and in the interest of privacy can be safely removed from further processing.

We use Relational Graph Convolutional Networks (RGCN) \cite{schlichtkrull2018rgcn} for the citation link prediction and case similarity tasks. \cite{muller2020integrated} presented an approach to use Graph Neural Networks for finding node similarity. \cite{ganesan2020link} discussed using link prediction from entities in a watchlist to very large graphs. \cite{shalghar2021document} exploits the document structure to improve RGCN performance. We can characterize these linked prediction methods as model based knowledge based construction.

\cite{chalkidis2020legal} introduced the LegalBERT model that continues to be used for tasks on the legal data including our experiments in this work. \cite{paul-2022-pretraining} have introduced InLegalBERT which is trained on Indian legal documents. Infusing knowledge into large language models has been discussed in several works. Two survey papers by \cite{wei2021knowledge} and \cite{yang2021survey} present different methods to infuse knowledge into large language models. \cite{islam2021fair} consumes a knowledge graph for the entity generation task.

KELM \cite{agarwal2020knowledge} proposed a method and datasets to convert knowelege graph triples into natural language sentences and further pre-train language models. SKILL \cite{moiseev2022skill} alleviates the need to convert triples to natural language and directly infuses knowledge into language models. \cite{kaur2022lm} proposes a methodical improvement to infusing knowledge into large language models.

\section{Legal Knowledge Graph}
\label{legalkg}

\begin{figure*}[!htb]
    \begin{center}
    \includegraphics[width=\textwidth]{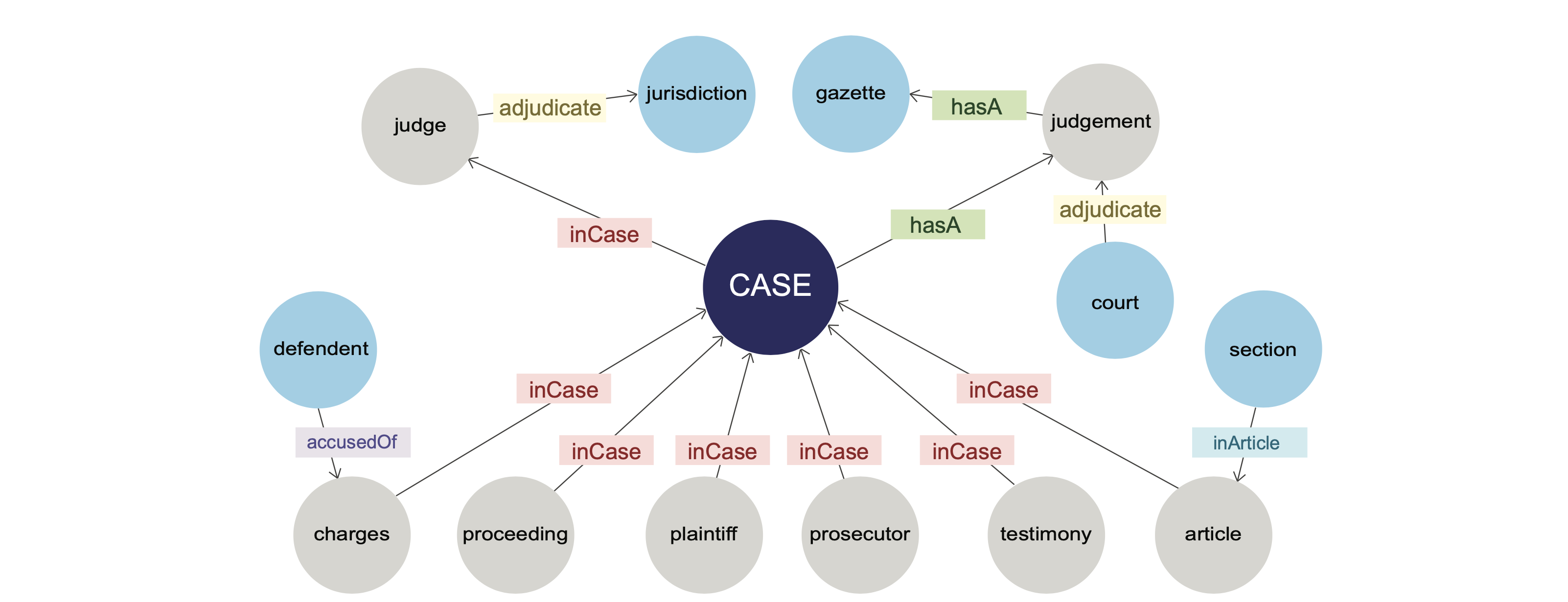}
    \caption{An ontology for legal documents}
    \label{ontology}        
    \end{center}
\end{figure*}

In order to create a legal knowledge graph, we began by scraping the web for court cases, judgements, laws and other cases cited from the judgements etc. In particular, we used the court repositories and other public sources in the Indian court system. For ground truth labels on the citation link prediction and case similarity tasks, we use the annotations from popular case search repositories, namely IndianKanoon \cite{ikanoon} and Casemine \cite{casemine}.

After selecting the above dataset of unstructured text documents, we construct a knowledge graph, similar to the methods described in \cite{agarwal2017cognitive} and \cite{vannur2021data}. For information retrieval comprising of document conversion, named entity recognition, fine grained entity classification and relation extraction, we utilize public NLP libraries like IBM's Data Discovery API \cite{shanmukha2019ddapi} and Stanza \cite{stanza}. To augment the off the shelf NLP pipelines, and to custom annotate entities and relationships, we use the SystemT annotators as described in \cite{chiticariu2010systemt}.

We represent the knowledge graph in triples format comprising of subject, object and predicate. The details of the legal knowledge graph populated by us are as shown in Table \ref{tab:akbc}.

\begin{table}[!htb]
\caption{Details of the legal knowledge graph}
\label{tab:akbc}
\centering
\begin{tabular}{lcr}
\hline
Documents     &{     } & 2,286   \\
Sentences    &{     } & 895,398 \\
Triples &{     } & 801,604 \\
Entities &{    } & 329,179 \\
Relations &{   } & 43 \\
\hline
\end{tabular}
\end{table}

Before we proceed further, we also derive an ontology for the our legal knowledge graph by a combination of manual process and topic modeling using Latent Dirichlet Allocation (LDA). We trained a topic model to surface the most relevant topics for the corpus and then handpicked some of the topics to construct the ontology shown in Figure \ref{ontology}.

\section{Case Graph}
\label{sec:case_graph}

Having constructed a legal knowledge graph as described in Section \ref{legalkg}, we now describe our method to create a case graph with a subset of information from the knowledge graph. Our knowledge graph consists of several node types like cases, laws, people, organizations and relationships between them.

For the case graph, we have used only a handful of node types derived from the ontology for the legal knowledge graph. With citation and similarity as two possible relations among the cases, first used the IndianKanoon \cite{ikanoon} search engine API to obtain IPR case citations. In order to further increase the size of the graph, we obtained the citation network for all the new documents obtained in the first batch through a breadth first search traversal on the document metadata. Ground truth for constructing the similarity network has been obtained from Casemine \cite{casemine}.

\begin{figure*}[!htb]
    \centering
    \includegraphics[width=0.9\linewidth]{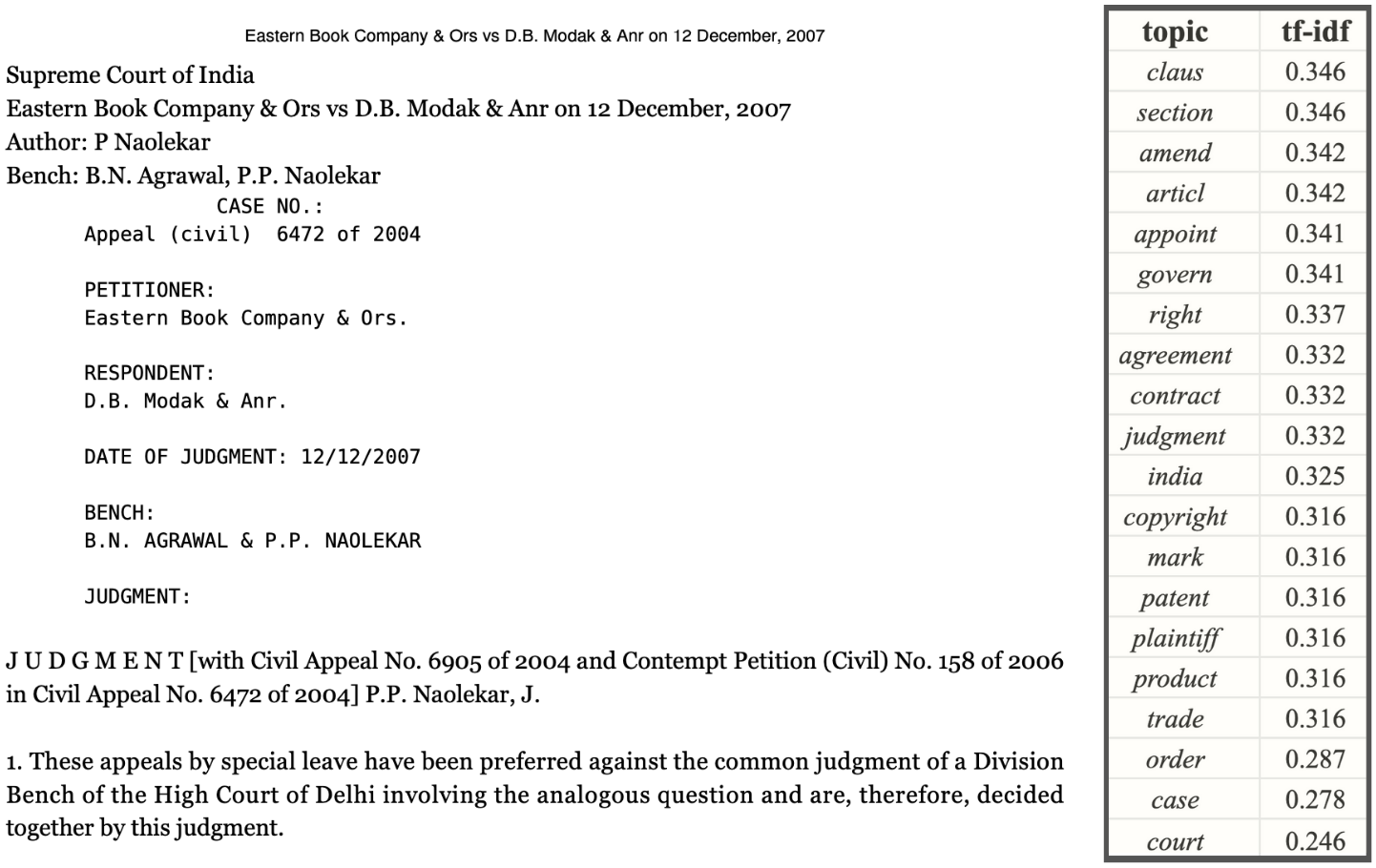}
    \caption{Topic modeling can help identify features for training GNN models on the case graph}
    \label{tfidf}
\end{figure*}

\subsection*{Feature selection}
\label{sec:feature_selection}

We obtained the important characteristics for each case using a combination of information extraction tools and domain specific annotation of the corpus. Given that cases hail from various number of public records, the structural variance in content layout is very high. The digitisation of old cases also leaves a scope of noise creeping into the dataset. In order to effectively extract information from such varied forms of digital legal documents, we use methods from \cite{chiticariu2010systemt}, \cite{ganesan2020anu} and \cite{stanza} for baseline information extraction.

\begin{table}[]
\label{lda}
\caption{Distribution of most significant LDA topics}
\centering
\begin{tabular}{|c|c|c|}
\hline
\textbf{Topic} & \textbf{\begin{tabular}[c]{@{}c@{}}Ontology concept\end{tabular}}  & \textbf{Occurrence}\\ \hline
compani                   & \begin{tabular}[c]{@{}c@{}}appellant,\\ defendant\end{tabular}   & 18.58\%     \\ \hline
court                      & court                                                              & 16.75\%  \\ \hline
right                     & --                                                                 & 13.33\%   \\ \hline
case                        & \begin{tabular}[c]{@{}c@{}}case,\\ judgement\end{tabular}          & 9.51\%  \\ \hline
plaintiff                   & plaintiff                                                         & 9.39\%   \\ \hline
section                     & section                                                          & 7.66\%    \\ \hline
servic                    & jurisdiction                                                             & 6.87\%       \\ \hline
govern                      & judge                                                            & 6.53\%  \\ \hline
state                      & --                                                                  & 5.70\%  \\ \hline
articl                     & article                                                            & 5.62\%   \\ \hline
\end{tabular}
\end{table}

For capturing the abstract subject matter of the corpus, we implemented the  Latent Dirichlet Allocation (LDA) topic model and computed the importance of the results using the TF-IDF metric. A sample excerpt from one of the cases with its topics is shown in Figure \ref{tfidf}. As can be observed from this example, topic modeling helped identify frequently used legal terms such as \textit{plaintiff}, \textit{section}, \textit{court}, \textit{agreement} among others, alongside a few concepts pertinent to IPR such as \textit{patent} and \textit{copyright}. These topics elucidated the foundation for a domain specific ontology elaborating on the relations that the entities share with each other. The knowledge graph constructed has the ontology as shown in Figure \ref{ontology}.

Finally, each of the domain-specific keywords and phrases were given a label from a selected number of broader and prominent IPR concepts based on their semantic similarities. These concepts were identified by legal experts in the domain. For instance, \textit{universal copyright convention} is extracted and labelled as textit{copyright} while law points such as \textit{patent cooperation treaty} and \textit{ghost mark} are labelled as \textit{patent} and \textit{trademark} respectively. We call these as law points and use as handcrafted features over and above the features automatically selected using the ontology.

\section{Experiments}
\label{sec:experiments}

We conduct our experiments on 2286 legal documents described in Table \ref{tab:akbc}. We used a train-test split ratio of 85:15 to conduct all the experiments and record our observations. For the citation link prediction task, the citations are present in the documents as available on IndianKanoon \cite{ikanoon}. For the case similarity task, two documents are similar if they are returned as similar cases by the website Casemine \cite{casemine} which we believe is partly annotated by experts. More details on this process are described in Section \ref{sec:case_graph}.

We started with a vanilla version of the RGCN model. We identified node features using the method described in Section \ref{sec:feature_selection}. This is our baseline model. We then added law points identified by legal experts as handcrafted features to our model. To populate these features, we annotated each judgement as containing a law point or otherwise, thereby yielding a model trained over 27 features as our first solution. As a separate experiment, we also encoded the node features using LegalBERT to leverage the contextual information in the documents into creating better feature representations.

We conducted our experiments using the open source Deep Graph Library \cite{wang2019dgl} on a single GPU. We use a typical train-test split in graph neural network models by removing 10\% of the edges from the case graphs and making the model predict the edges back. We use same number of non-existing edges as negative samples. We trained the vanilla RGCN model for 1200 epochs on the citation link prediction model and 600 epochs on the case similarity model. For the models with law points as features, we trained them for 400 epochs.

The performance of the RGCN model for \textit{citation} and \textit{similarity} tasks are as shown in Table \ref{results}. As can be observed from the ROC-AUC scores, the model with handcrafted features performed better than the vanilla version in both tasks, returning a score of 0.620 for citation prediction and 0.556 for case similarity. Encoding the features with LegalBERT considerably improves the performance on the citation prediction task but has negligible impact on the case similarity task. We discuss why this might be happening in Section \ref{sec:knowledge_infusion}.

\begin{table}[htb]
\caption{Performance of RGCN models using ROC-AUC scores}
\label{results}
\centering
\begin{tabular}{lccc}
\hline
\textbf{Model} & \textbf{Citation Prediction} & \textbf{Case Similarity} \\
\hline
RGCN baseline     & 0.587                & 0.513           \\
RGCN + handcrafted features   & 0.620       & \textbf{0.556}   \\ 
RGCN + LegalBERT  & \textbf{0.725} & 0.550 \\
\hline
\end{tabular}
\end{table}

In order to explain the predictions, we juxtapose two similar cases and compare their underlying features as shown in Figure \ref{fig:node_comparison}. The visualization includes multiple subsets of the predicted network and displays information for two selected cases at any instance. This information comprises of the metadata and the feature set values consumed by the RGCN model during the training process.

\section{Discussion}
\label{sec:discussion}

In this section, we discuss potential alternatives to our solution. Treating the case similarity problem as a graph link prediction task enables us to focus on important features rather than the document structure and language constructs of the legal documents. However, case similarity can be treated as a document clustering problem too.

\subsection{Unsupervised approach to case similarity}
In an attempt to replicate the RGCN case similarity model results and its AUC scores recorded in the experiments, we adapted to a non-neural unsupervised methodology of density based clustering. By implementing DBSCAN over the similarity network defined by the feature set used in link prediction, we aim to maximize the intersection of the similar case prediction results obtained from these two disjoint outlooks.

While neural models tend to benefit from high-dimensional data to a certain degree, they had to be reduced significantly using Principal Component Analysis in order to define a set of clusters using this feature set. The cluster formation as summarized in Table \ref{tab:cluster_details} has been carried out on the document level, wherein each node represents a case from the similarity network. Of the 27 handcrafted features in Table \ref{results}, we used 18 for creating the principal components for DBSCAN clustering.

\begin{table}[!htb]
    \caption{Details of our study on text clustering for case similarity}
    \label{tab:temps}
    \begin{subtable}[t]{0.5\textwidth}
    \centering
    \begin{tabularx}{\textwidth}{Xr} 
    \hline
    Clusters    & 2   \\
    Neighborhood Radius     & 0.44 \\
    Principle Components     & 4 \\
    Average Intra-Cluster Edges    & 439.5 \\
    Average Inter-Cluster Edges & 406 \\
    Average Coverage & 0.3396 \\
    \hline
    \end{tabularx}
    \caption{Details of DBSCAN Clusters }
    \label{tab:cluster_details}
    \end{subtable}%
    \hfill%
    \begin{subtable}[t]{0.5\textwidth}
    \centering
    \begin{tabularx}{\textwidth}{Xrr} 
    \hline
    \textbf{}   & \textbf{Cluster 1}             & \textbf{Cluster 2} \\
    \\
    Test Samples      & 74                & 62          \\
    \\
    RMSE    & 0.5312       & 0.3422   \\ 
    \\
    \hline
    \end{tabularx}
    \caption{RMSE Results}
    \label{rmse}
    \end{subtable}
\end{table}

As can be observed from the clustering results recorded in Table \ref{tab:cluster_details}, the density-based approach did not form an appreciable count of distinct clusters for the considered feature set and tuning parameters. Moreover, the average inter-cluster edges has a comparable value to the average intra-cluster edges, thus implying towards a scope of improving the clustering inputs and parameters.

\begin{figure}[htb]
    \centering
    \includegraphics[width=\columnwidth]{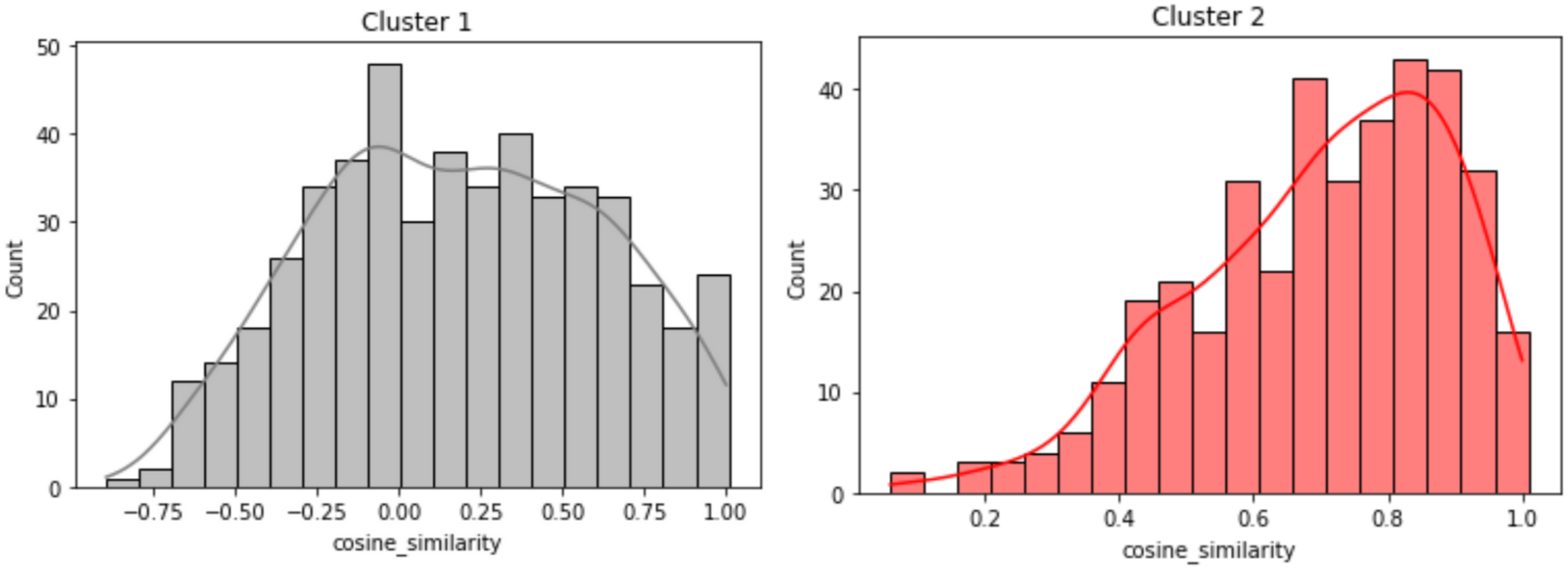}
    \caption{Cluster-wise cosine similarity Distribution}
    \label{fig:clusters_cosine}
\end{figure}

The cluster 1 cosine similarity distribution in Figure \ref{fig:clusters_cosine} indicates towards a potentially better partitioning of the nodes and to consider only the similar nodes within a single cluster. Table \ref{rmse} specifies the cluster-wise RMSE scores obtained over the test samples from the RGCN experiment. Assuming the AUC scores of these samples to be the true values, the cosine similarity score is computed for all test set pairs of nodes within the same cluster in order to compute the error.

From the RMSE scores recorded in Table \ref{rmse} and the similarity distribution scores from Figure \ref{fig:clusters_cosine}, it can be inferred that the only 2 clusters that are obtained through DBSCAN fail to effectively capture the overall characteristics of our dataset, thus forming a weaker basis for predicting similar cases and identifying missing links. In comparison to the neural model approach, this unsupervised alternative to case similarity computation lacks the capacity to handle large dimensions of data, and thus leaves a scope of curating a better feature set specific to clustering analysis of cases.

\subsection{Legal Knowledge Infusion}
\label{sec:knowledge_infusion}

Another way to look at our results in Table \ref{results} is the impact of domain specific terms in the model predictions. One of the reasons for the relative poor performance of our case similarity models in general, and in particular while using LegalBERT could be the absence of Indian IPR law terms and terms used in Indian court judgements in LegalBERT. While we have tried to incorporate terms specific to Indian IPR (Intellectual Property Rights) judgements as separate features in our RGCN implementation, they are not exhaustive.

A better approach could be to infuse domain specific (IPR) terms into a large language model and use such an LLM for encoding node features. We could use InLegalBERT to overcome this problem to some extent, but this will not generalize to legal documents from other countries. We however, hope to explore the use of InLegalBERT in the future. As described in Section \ref{sec:related_work}, there are several methods to infuse knowledge into even bigger autoregressive large language models. KELM \cite{agarwal2020knowledge} generates natural language sentences from knowledge graph triples before using them for training a large language model. SKILL \cite{moiseev2022skill} provides an elegant way to directly infuse knowledge graph triples into LLMs.

We also observe that using LegalBERT or any legal knowledge infused large language model will face some of the limitations faced by us while using LegalBERT. One of the ways to overcome this problem could be to separate LLMs and the external knowledge source, namely the legal knoweldge graph in our case. \cite{kaur2022lm} proposed this separation as a way to enable continuous updates to the knowledge graph while avoiding the need to avoid repeated pre-training or fine tuning of the large language models.

\section{Deployment}
\label{sec:deployment}

Our case similarity recommendation system is deployed as a cloud foundry application on the IBM Cloud. But it can be deployed on a stand alone server or used as a desktop application. Users of our system can select a case from a list or use a search engine to find a selected document. When a case is selected, a graph containing the case, it's citations and similar cases is displayed. The edges of this graph are based on the predictions from our GNN models. Newer cases can be added to the system and links predicted using our models, though currently only batch mode predictions are supported for newer cases.

\begin{figure*}[htb]
    \centering
    \includegraphics[width=\textwidth]{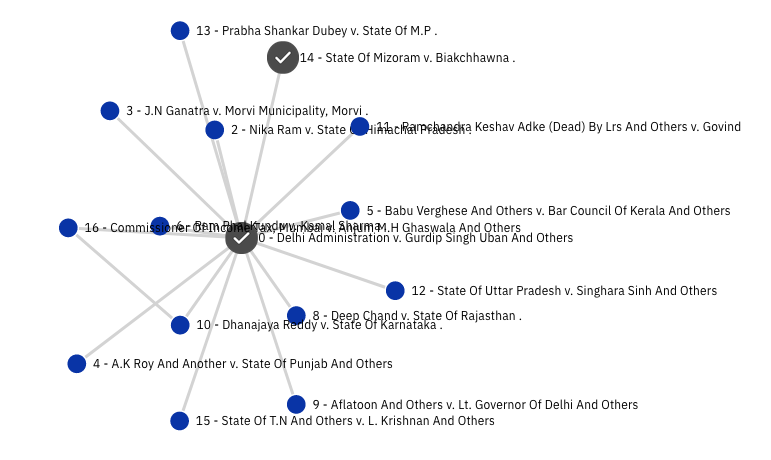}
    \caption{A subgraph with predicted links}
    \label{fig:example_graph}
\end{figure*}

\begin{figure*}[htb]
    \centering
    \includegraphics[width=\textwidth]{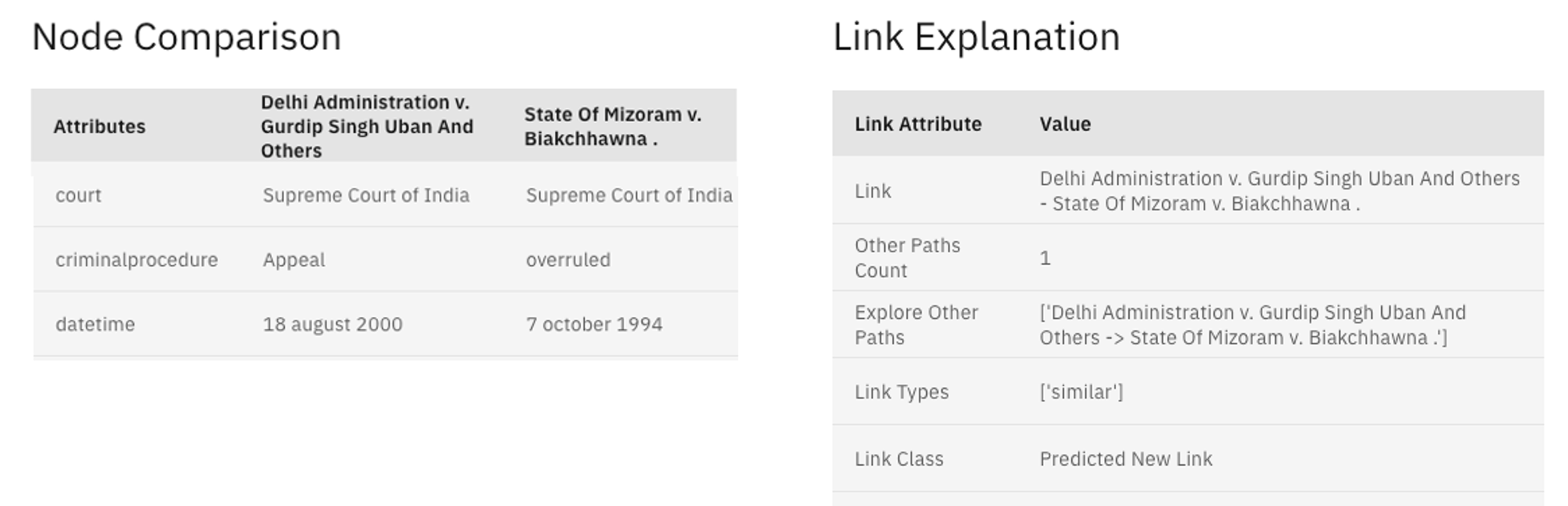}
    \caption{Comparison of nodes and explanations for the predicted link}
    \label{fig:node_comparison}
\end{figure*}

Users can click on edge to learn how the two cases are related to each other. We showcase important features of the cases. While currently these are all features used in our experiments, we could use GNN Explainability techniques to identify node features and edges that were most responsible for a prediction. We leave this as feature work. As shown in Figure \ref{fig:node_comparison}, users of our system can identify similar cases from past judgements and also see the verdict, and important law points on which the cases were decided.

We instead explain each prediction by show-casing the most prominent path between two similar cases, as shown in Figure \ref{fig:node_comparison}. It is likely that two similar cases have common judgements, or legal documents, if not directly linked to each by a citation link. By using a path ranking based approach as described in \cite{ganesan2020link}, we help the end users understand why a case is recommended to them. For the citation prediction task, we can optionally display the citation text in a judgement.

In the future versions of our implementation, we intend to use GNN model embeddings to improve search on the case documents. We can also use our legal knowledge graph to enable faceted and semantic search capabilities on our solution. We leave these for future work.

\section{Conclusion}
We constructed a legal knowledge graph from a corpus of Indian judicial and legal documents, predominantly focusing on intellectual property rights cases and related legislation. We trained graph neural network (GNN) models for citation linked prediction and case similarity. We showed that incorporating domain relevant features extracted using topic modeling and inputs from domain experts results in improved performance. Such knowledge graphs can also be used to infuse knowledge into or fine tune large language models (LLMs) before using them in downstream tasks.

\begin{acknowledgments}
This work was done partly under IBM's Global Remote Mentorship Program. We thank Mona Bharadwaj and Poornima Iyengar for their support. We thank the reviewers for their feedback.
\end{acknowledgments}

\bibliography{sail2023}




\end{document}